# Szloca: towards a framework for full 3D tracking through a single camera in context of interactive arts.

Sahaj Garg

**Abstract**— Realtime virtual data of objects and human presence in a large area holds a valuable key in enabling many experiences and applications in various industries and with exponential rise in the technological development of artificial intelligence – computer vision has expanded the possibilities of tracking and classifying things through just video inputs, which is also surpassing the limitations of most popular and common hardware setups known traditionally to detect human pose and position, such as low field of view and limited tracking capacity. The benefits of using computer vision in application development is large as it augments traditional input sources (like video streams) and can be integrated in many environments and platforms. In the context of new media interactive arts, based on physical movements and expanding over large areas or gallaries, this research presents a novel way and a framework towards obtaining data and virtual representation of objects/people - such as three-dimensional positions, skeltons/pose and masks from a single rgb camera. Looking at the state of art through some recent developments and building on prior research in the field of computer vision, the paper also proposes an original method to obtain three dimensional position data from monocular images, the model does not rely on complex training of computer vision systems but combines prior computer vision research and adds a capacity to represent z depth, ie- to represent a world position in 3 axis from a 2d input source.

———————————— ◆ ————————————

## 1 INTRODUCTION

T HE data of existence and position(in 3D) of objects, (and furthermore of people – in context of this research) in an area enables and unlocks potential use cases for many applications and industries including AR/VR/Mixed Reality experiences, Gaming, Immersive Interactive experiences, Performance Arts, Motion Capture, Training, Human-Computer-Interaction, Surveillance, Human Behavior understanding, Pedestrian- Crowd Understanding/Transportation, Activity Recognition , Object Inventory Tracking etc.

New Media Art is rapidly evolving alongside technology and the possibilities to detect changes caused by human presence in an environment has opened up a new era of ways to interact with physical surroundings in integration with technology and create physical movement-based art forms and games.

The framework of this research is towards the goal of detecting multiple objects or human presence in an area along with their pose and location data, in three-dimensional space, providing us a virtual representation of these objects / people. We look at ways to obtain 3D world space position, 3D pose/skeleton, tracking/ID and masking the detections. The research aims to integrate the function in more traditional setups, ie- from a single camera (monocular images/videos) instead of using systems such as lidar's, depth sensors or putting any hardware on the players/audience. This enables the experiences to be more flexible, deployable and accessible to a wide array of users and locations. The framework surpasses the limitations of famous sensors such as kinect, realsense - like low field of view/range and limited capacity of number of skeletons which can be tracked, while also making it cost effieicient.

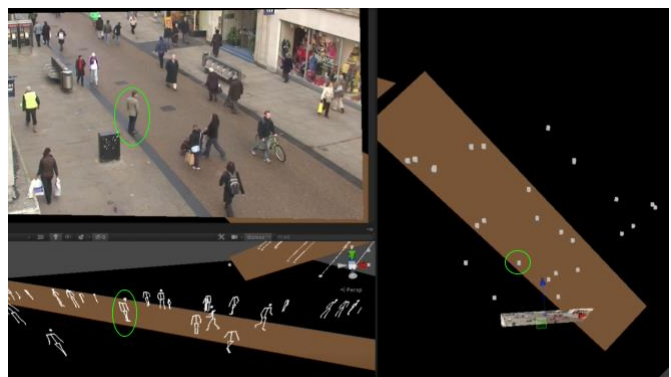

Fig. 1: top left - image input, right – top view of obtained three dimensional positions where white dots represents people and the brown rectangle represents pathway for reference as seen in the image input, bottom left – skeletons in three-dimensional space.

The underlying foundation of this research relies on some of the recent developments in the field of computer vision inclining towards human pose detection and object tracking which utilize deep convolutional neural networks (CNNs). While we do not discuss the deep workings of these neural nets, looking at the state of the art, we discuss their limitations and how they can be tangled together to

• Sahaj Garg. E-mail: sz.sahaj@embracingearth.space

overcome them. We also dive into some of the problems and challenges in the field of detection over large spaces and suggest an original method in support with these computer vision algorithms to convert 2D detections into 3D position. Few recently published researches [50] [9] [46] are able to provide 3D position of multiple person with single camera, however lacks tests over large areas/fields, szloca's approach to obtain 3D position may perform much better in these situations and with a better frame rate for large number of objects/people. Our final aim is also inclined towards development in more artistic friendly environments (such as Unity3D game Engine). This paper can also be followed up as a quick inventory and state of the art in the field of computer vision inclining towards human presence and tracking.

The initial exploration of this research is conditioned towards the desire to create interactive and immersive movement-based games, art installations and performances in three-dimensional environments that occupy an entire galary or large spaces.

## 2 BACKGROUND

Arts and games industry evolve rapidly with new Technology [1] and goes parallely in collaboration with the technological advancements and is often amongst first pioneers in experimenting and innovating. There are many artists and interactive works emerged which include some form of technologies and human interaction system, referred under the umbrella of New Media Arts [2]. The term New Media Art incorporates Art forms which make use of technology to create itself. The term has evolved and risen alongside the technological innovation from 19th century.

Frank Popper in Technological to Virtual Art [36] traces the creation and rise of immersive new media art forms in conjunction with technology from 19th to early 20th century. He defines the evolving new media art forms which include movement-based interactions alongside vision and hearing in a term as 'immersive virual reality' because of the intensity of experience and expression these art forms offer. Popper identifies new media Art forms and the very important role technological advancement in enabling experiences undreamed of before. In Art, Action and Partcipation [35], Popper further elaborates on how movement-based art forms – kinetic art plays an important factor to involve and immerse the spectator both actively and totally in the event and refers it as progressive liberation of the spectator from mere participation to creative action.

Oliver Grau [37] in Virtual Art From Illusion to Immersion emhpasises on how immersion plays a key factor in the development of art forms and that it can be a very intellectually stimulating process which holds the power to change or create a passage from one mental state to another. He emphasizes that the new art media with aid of technological advancements encompasses a potential for high grade virtual immersion with the basis of interaction and stimulation of other senses as well instead of just eyes which may include the body and movement of viewers in three-dimensional space, which can also be said to be in relatedness with Paul Dourish [38] in 'Where the Action Is: The Foundations of Embodied Interaction' where he suggests the term physical embodiemrent to define how the action and play between human-computer interaction systems happens in the physical space rather than just on screen, and merging the systems beyond screens into physical is an important and evolving niche to be looking at.

Myron Kruger [36] was one of the early Artist who started experimenting with interactivity through body movements in Art installations. His work 'VideoPlace', developed in 1975 used tech to display 2D silloutte of its viewer in a collaborative art installation. The installation was one of the first of its kind interactive art which uses human body masked as a 2D silloutte on screen which was able to interact with on screen/projected virtual elements.

Body movement-based Art holds unprecedented ways of engaging users, it catches instant reaction and embodies one into their physicality. These forms of art encourage collaboration, joy, freedom, expression and play. An example of which is seen in this video documentation by Carte Noire [45].

Following in 20th century, a lot of new media art have emerged which uses body and movement of its viewers as a means of interaction and creation. Some of the New Media Artists and Collectives experimenting around these ideas include Artechouse [41], Light Art Space [42], Teamlab [58 ], Adrien M & Claire B [44], FUSE Interactive [59], Chunky Move [57], Refik Anadol [60], Responve-Simon Burgin [52], Andrew Quinn [63], Asterix – Peter Walker [64], Kuflex [62] , Craig Walsh [53] , Ecoscreen [54], PolyMorphArt [55], Design IO [61], Daito Manabe [64] and more.

'Hana Fubuki' is a modern-day interactive Art Installation by Akiko Yamashita [40] in a full gallary setting which uses players full body movement for interaction. She explains how the installation uses 6 kinect sensors to create a seamless whole and is divided into individual playspaces. These individual playspaces are divided by range of detection with gaps in between and are also limited to maximum number of people which can be detected by these sensors in each space. Akiko emphasizes on this while mentionning how sometimes players when in groups wont notice and keep trying to interact even if they are not detected.

The Treachery of Sanctuary by Chris Milk [43] powered by Creators Project – Art and Culture by VICE, plays with the idea of transformation of its audience sillouttes as shadows. As the creators explain, one of the technical barriers they hit was getting crisp masks of people for shadows. The hardware used in the project is a Kinect sensor, through which they get skeletons of a person, however they were unable to implement crisp structures of peoples silhouettes, and settled towards a more artsy jagged outline which was not the first intention.

In the paper 'Dancing with Interactive Art's [56], the authors describe a hybrid mix of using computer vision with Kinect Sensor to detect world position and create touch interactivity in an interactive dance performce. A similar redition of using Kinect sensors is seen in Dokk by Fuse Interactive [59]. The framework of this paper offers alternatives while expanding on the technical possibilities which also addresses the inheritent limitations and cost efficiency of traditional sensors.

## 2.1 Technological advancement

The growth in advancements of artificaial intelligence has been on exponential rise, computer vision is a sub category of the field which looks at training computers to interpret and understand the visual world by using digital images from cameras and videos and deep learning models, machines can accurately identify and classify objects and react to what they see. Computer vision has its applications among a range of industries and Arts sector is not so behind in using and experimenting with the tech, we have seen a significant boost in Augmented Reality Art experiences which relies on computer vision to overlay on real world objects.

Traditionally, one of the common and very first thoughts which comes to mind when talking about human and pose tracking is using depth sensors [48], which use invisible near-infrared light and measures its "time of flight" after it reflects off the objects, such as Kinect, Realsense, Xtion etc. These sensors are limited by their relatively very small field of view and a very low number of maximum skeletons which can be detected. Most of these commercial depth senors are not made for harsh weather conditions which means they cannot be used in open spaces and are also relatively expensive.

One of the other methods used in understanding the physical environment virtually is by using Lidar sensors. Lidar works on similar principal as depth senors but uses lasers for longer detection and generates a point cloud of the surface which can be further processed to identify and classify objects. However, this grows exponentially both in terms of cost and computation power required and is a very expensive option, also a fact, why Elon Musk switched to aid of computer vision[49] for his auto vehicle company Tesla contrary to lidar systems which were famous in the industry

Using recent advancements in artificial intelligence - computer vision and frameworks defined in this paper, we can achieve results in a very cost-effective way through just a video/ single camera feed. This makes the experiences accesible instantly to a lot of people and locations both indoors and outdoors as it also can be extremely weather resistant just by using a weather resistant camera.

But the main supremacy this system has is its field of view,

range of detection and capacity of number of people/objects it can track. It can very easily cover the playarea which would have initially required multiple sensors - expensive hardware for integration. The field of view can be extended just by a switch of camera's lens or zooming, and it can also be placed further to cover the whole area of detection. The method we propose can get 3D positions of objects/people even at a very large distant. The number of objects and people which can be tracked is infinite depending on gpu/computer the framework is running on as compared to often 4-6 people at max in most traditional sensor-based solutions.

## 2.2 Authors Projects

Cosmic Revelation (name tbd), is an interactive art installation with Orcha Collective, initially to be commissioned for Moonlight Festival 2020 at Bunjil Place, Melbourne. The idea and concept of the installation revolves around detecting interactions and movements of the players with creative projections and digital screens for display. The playarea spans over a large open field space of 20x20sq meter open space which is filled with people. Tracking of a lot of people/objects in three-dimensional form over such area becomes relatively harder, expensive, limiting to number of players and interfering in between the open playspace when using traditional/common methods of sensors. The methods discussed in this paper aids to that and overcomes limitations by using a single well-placed Camera.

Brunswick Ball is an interactive installation combining dance and music. To be showcased outside Brunsick Library which has been intertwined with its former use as a Ballroom. The installation features interactive touch surfaces on rocks and physical surroundings as a means to collaborate and create music and trigger visuals, and captures it's players and viewers movements in beautiful sillouttes projected on the windows of the Library, as if one could have seen when the Library was a Ballroom in 1980's. The methods in this paper aids to live movement visualization and interaction. Brunswick Ball by Sahaj Garg (Author) and Carlo Tolentino is part of a bigger installation project Playable City [47] – Melbourne Edition, with creative director Dr Troy Innocent [67] and is commissioned by the City of Moreland - council of Melbourne. Playable City aims to encourage play in urban environments in integration with technology and connect elements that already situate in the city.

Author is further experimenting and creating prototypes and installation projects which uses people-pose / object tracking in three-dimensional physical space to create simulations, physical games, live theatricals, performances, immersive spaces, art installation, tracking solutions and more [69].

## 3 INVENTORY

In this section we look at some of the available and related toolsets and research in the field of computer vision which aid to our method. Most of these researches exists outside game engine or editor based environment but lays the ground foundation. We use some of these toolsets and build upon, which is discussed in the methods section. We also discuss their limitations and ways on how these initial understanding can be tangled together and used with each other to overcome their limitations and create a full system.

### 3.1 Detection and Tracking

Yolo: You Only Look Once: Unified, Real-Time Object Detection [16] [15] [18] is one of the most famous algorithms based on a single convolutional network trained to detect objects in an image. Yolo does not provide pose or masks or skeleton tracking but provides 2D bounding boxes around detected objects.

Simple Online and Realtime Tracking (SORT) [3] [13] is a pragmatic approach to multiple objects tracking with a focus on simple, effective algorithms. This network trained on deep learning data sets does not give pose however tracks the objects identity and keeps it same during the time its in the frame (in 2D).

Detect and Track [25] Published via facebookreseacrh, detects human pose skeleton as well as can maintain ID of them, however, is 2D based.

### 3.2 Mask and Skinning

DensePose: Dense Human Pose Estimation in The Wild [4] - published through facebookresearch provides surface representation of the Human Body(ie-mask), is 2D based both in sense of position and pose.

VIBE: Video Inference for Human Body Pose and Shape Estimation [CVPR-2020] [10] Provides 3D human Pose as well as shape (thickness of skeleton), however not the 3D position/camera coordinates.

Detectron [26] Published via facebookresearch, detectron is a Mask R-CNN implementation which provides masks over detected objects in a video.

### 3.3 Pose/People Detection

Lifting from the Deep: Convolutional 3D Pose Estimation from a Single Image [5] uses multi-stage CNN architecture, building on top of 2D pose recognition and provides 3D pose, however, is intended for single person detections and doesn't provide position in 3D.

OpenPose: Realtime Multi-Person 2D Pose Estimation using Part Affinity Fields [19] provides pose estimation in 2D, can also be used to detect object/vehicle keypoints.

Integral Human Pose Regression [11] aims to provide 3D pose (tested on single person images), does not provide 3D position.

3d-pose-baseline [14] provides 3D pose of single person cases only, without including 3D position.

VideoPose3D [12] published by facebookresearch provide 3D pose – 3D position, however, is aimed towards single person detection.

Onsensus-based Optimization for 3D Human Pose Estimation in Camera Coordinates [8] Provides 3D pose of a person however uses multi camera view to provide more precision for position coordinates.

Fast and Robust Multi-Person 3D Pose Estimation from Multiple Views [68] and Multi-person 3D Pose Estimation and Tracking in Sports [66] can provide 3D pose estimation alongside 3D position of detected person however requires 3-6 cameras from different angles to work.

Depth-Based 3D Hand Pose Estimation: From Current Achievements to Future Goals [6] looks at the state of the art in hand-pose detection using CNN's - from monocular images.

*Closest to providing 3D pose and 3D position (skeleton) of multiple people from single camera, lacks data for detection over large fields, lacks integeration with unity (for now) –*

XNect: Real-time Multi-Person 3D Motion Capture with a Single RGB Camera [50] can provide 3D skeletons and 3D camera positions.

Multi-Person 3D Human Pose Estimation from Monocular Images [9] brings out the goodness of CNN providing 3D pose as well as approximate global position coordinates for multiple people in the frame.

Camera Distance-aware Top-down Approach for 3D Multi-person Pose Estimation from a Single RGB Image [46] provides 3D multi-person pose estimation from a single RGB image and global world positioning as well without the need of any ground truth.

### 3.4 Depth Estimation

Video Depth Estimation by Fusing Flow-to-Depth Proposals [7] , Deeper Depth Prediction with Fully Convolutional Residual Networks[27], Consistent Video Depth Estimation [34], and Deep Ordinal Regression Network for Monocular Depth Estimation [33] aims towards providing video depth estimation from a single camera - monocular sequences.

## 4 METHOD

### 4.1 From 2D screen position to 3D world position, getting z-depth

One of the main components and original findings of 'szloca' is converting 2D detected screen scpace positions into 3D space, which can be easily integrated in game engines and other editor-based artist friendly software environments and is able to provide three dimensional positions over larger areas.

The method proposed works on shooting raycasts from camera's location, angled towards the objects screen-space position- which is obtained by object detection algorithms. The raycast is further allowed to travel infinitely till it hits a virtually aligned terrain/ground. The world position of the collision of raycast to the aligned ground gives us the z axis, relative to x,y positions in the frame which was obtained from initial object detection.

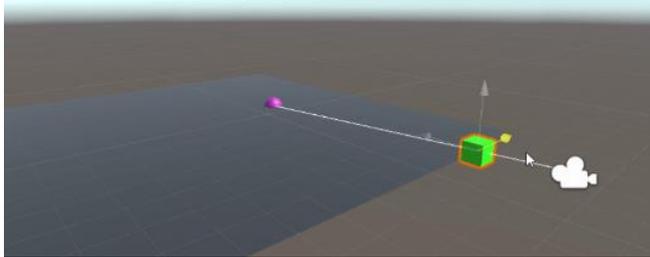

Fig. 2: Basic workings of szloca's method of converting 2D to 3D, green cube representng the detected object (in screen space), The camera shoots a ray aligned towards it till it hits(pink) the ground terrain(grey).

It is important to note here - to experiment with the perspective and orthographic settings of the virtual camera, depending on the image and camera placement. With a perspective camera, the raycasts will be affected by camera's zoom angle, giving inaccurate distances farther the object is in the video, hence in some cases orthographic camera is better choice however, this widening of ray angles when using perspective camera, can be used to compensate for the the visual angle which become less as the distance increases making distance between objects looking smaller than they are.

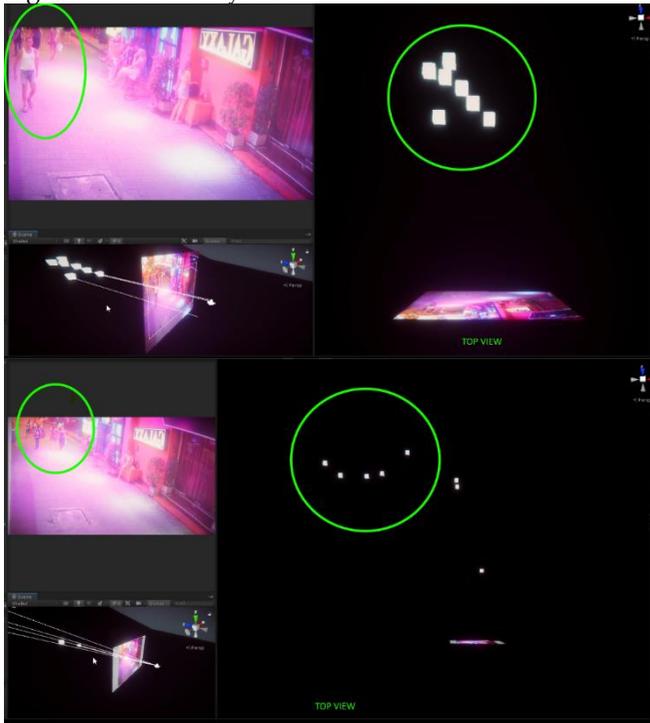

Fig. 3(top) & 4(bottom): Showing difference between orthrographic and perspective camera systems. On left side of images is input video and on right side shows the Top view for positions from our system. The top image

represents orthographic system to detect and shoot rays, here rays are not affected by any angle but are shot in straight 90 degrees, as we can see the farther people are in the video we start to see them crowded up in our virtual output, which is not right in real world positions. Bottom image shows using perspective camera to shoot rays and the angle distortion of rays to compensate for the visual angle, which provides us better position values in terms of real world.

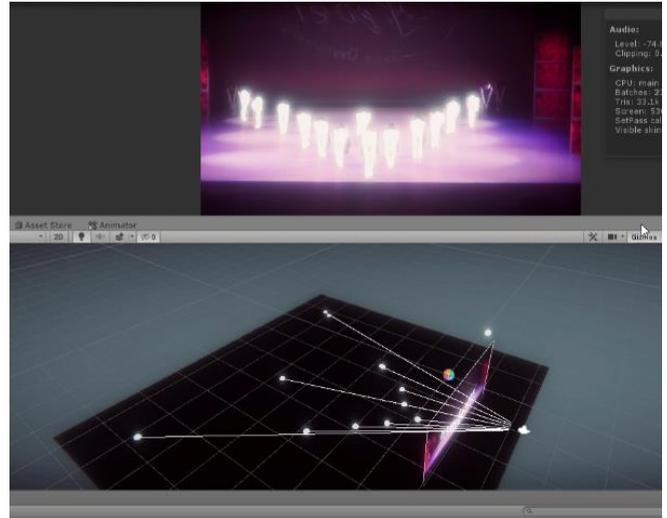

Fig. 5: Showing the initial workings of getting 3D position data from 2D screen space. Built on screen-space human pose detection by OpenPose algorithm. At the top of the image is video feed with people standing in a V formation on a stage. At bottom of the image, the working of szlocas's ray system and obtained 3D position is shown.

It is also important to choose which part of the tracked skeleton the ray is assigned to pass through, as it initially depends on screen position of the object. Passing it through head/nose imposes problems with people of different heights, for example people with different heights walking parallel to each other will give inaccurate 3D world positioning of szloca if ray assigned towards head as the screen position of head/noses of these two differ in y-axis. The first solutions to solving this issue is assigning the ray to feet of the obtained skeleton, which seemed to work and deliver more accurate positioning in such cases, however when a person walks, the feet is often lifted from the ground this again gives jitters in z-depth as y-axis of object is changing irrelevantly. A better possible solution seems to be to pass the ray through waist or lower torso. To Note- The ground truth also needs to be fixed with some error correction here if aligning rays with torso to make up for the height of legs to torso. An effective method here we experiment with is calculating the approximate center of the person and attach it to a point below on ground and pass the ray through this point. Another solution is to use the bounding box provided through Tracking and ID section (discussed further) – ray aligned with lower frame of line in middle of the bounding box. This can further be smoothed using functions to avoid jitters in the obtained position of moving objects/people.

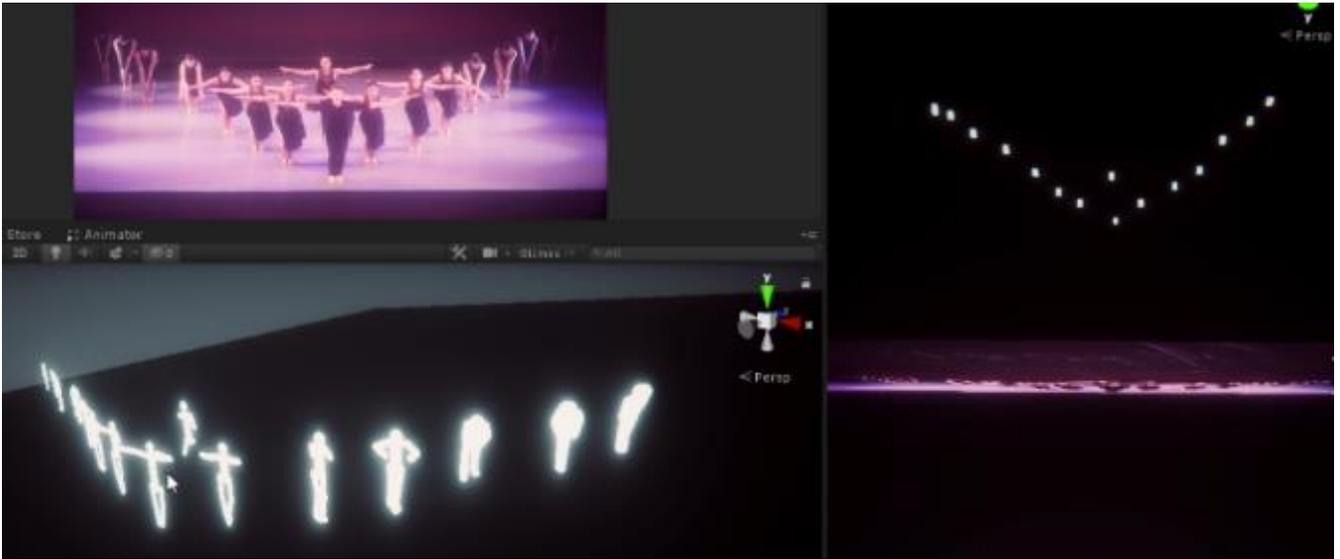

Fig. 6: Top left – input video, right – top view of positions obtained, bottom left – skeletons in 3D world space from a slight right-angled view.

The position of skeleton mask (obtained from Openpose Unity Plugin in this iteration) is further copied to their respective individual acquired z-position, giving us virtual skeletons in 3D world space. The height and size of the detected skeletons also gets smaller the farther they are in the video, to compensate for this, we can also scale the size of detected skeletons according to the obtained position data, i.e.- the farther they are from the camera, the more it is scaled up to compensate for the visual perspective distortion.

### 4.2 From 2D skeletons to 3D skeletons – 3D Pose estimation

In section 4.1 we've discussed ways to obtain 3D position, we further copied tracked skeletons into 3D positions, however the skeletons themselves are in 2D. i.e – they exist in a flat plane as they are initially detected from the video. Having just the full 3D position data of objects enables a lot of room to create, however this can further be refined if our virtual skeletons are in 3D as well instead 2D. This enables ability to play with more macro interactions between the users and the environment, such as obtaining exact data if players are touching a surface, their rotation values, more precision on how the players are moving their limbs in 3D space and more.

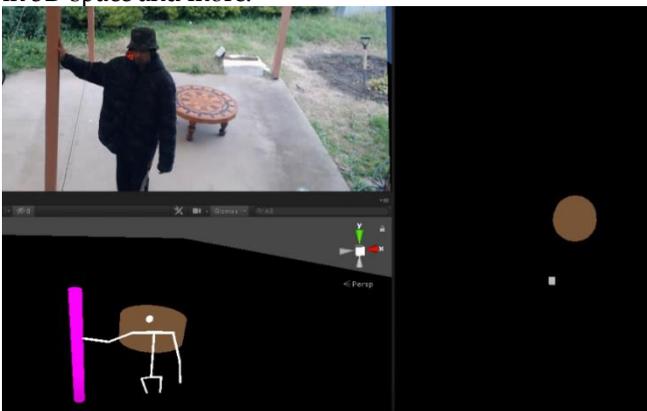

Fig. 7: Showing how the framework can be used to detect

interactions and make huge surfaces interactive without the need of any hardware on the surface. To note: The example in the image is built with 2D skeleton.

Getting 3D skeletons of multiple people from a single image is a challenging problem and is being experimented and researched on everyday. Openpose initially provides 2D posetracking, for example - if hands or body moves perpendicular to camera it'll still give skeleton in 2-dimension making the length of virtual skeleton/arm smaller, to resolve this issue and get full 3D skeleton tracking, it can be combined with [14], as seen in [20] - however limits that to single person detection. OpenPose also has in development 3D skeleton tracking however is limited to single person as well because it is unable to track the person identity [21]. To implement and scale this feature of 3D skeletons to multiple people, each section of the video of detected people (obtained and cropped by tracking and ID module 4.3) can be treated as individual videos, each containing single person and feeded into such. Doing so provides us with 3D skeletons of multiple people from a single image/camera feed - containing multiple people.

One of the other methods to do this can be seen in [22] where depth values are assigned to 2D skeleton joints to create 3D skeletons. However, in the example [22] a stereo camera -Zed by stereolabs is used to get depth values. In an attempt to attain everything through a single video/camera, Depth Estimation Algorithms [7] [27] [33] [34] can be integrated to achieve the depth through monocular images/video, these can also be used in lieu of szloca's ray method for 3D position, however is a much difficult implementation. Further experimentation and comparisons in different types of footages are needed to determine the pros and cons. Some of other recently published prestigious work along the lines include [50] [9] [46] [66].

### 4.3 Tracking and ID

All the methods to detect and attain the pose and position discussed until now works perfectly for single images and may look like they are working well with videos – which is basically a sequence of images. However, computation happening in each of these algorithms is frame based, meaning every skeleton pose detected every frame is unique on its own. When initially combining with szloca's raycast approach this became more relevant as one can see the jumping of containers in the background console as it goes on assigning unique skeletons every frame when there are multiple present in the scene.  For optimal use cases maintaining the ID of the person/object is essential, to recognize it as one entity during its time in the video frame

This problem is solved with DeepSort [3]. DeepSort integration with PoseEstimation and detection algorithm such as OpenPose or Yolo, can link current and previous frames enabling to maintain and keep the ID of objects, seen such as in [23] [24] [25].

Entangling, this tracking module with other module of position and pose detection also brings a lot of aid, accuracy and smoothness to position and pose detection.

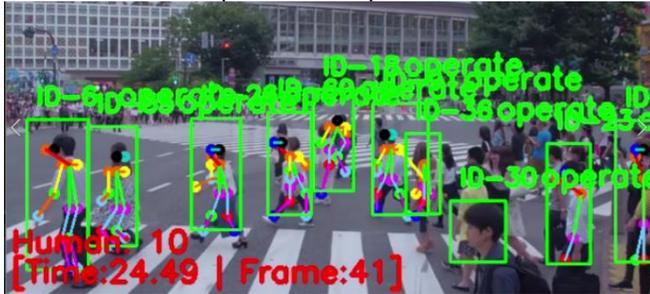

Fig. 8: Showing bounding box provided by algorithms such as yolo /deepsort and its integration with openpose for tracking id in [23].

### 4.4 Masking and Skinning

Having an accurate mask around the detected people/object can be helpful in many ways, such as rotoscopy, providing mass and getting exact shape of people/objects. Access to this layer of data enables a lot of room to create and expand, such as for Artists to experiment with artistic renderings and visualization and visual classification of the objects. This can be done in both 2D [4][10] [31] [26] and 3D [28] [29] [30].

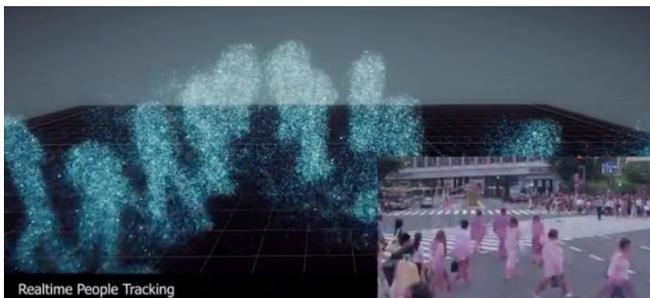

Fig. 9: Example of how stylised renders can be used with masking/detection for artistic endevours [51]. Using computer vision, we are able to do this for crowds instead of just a few people through sensor-based pipelines.

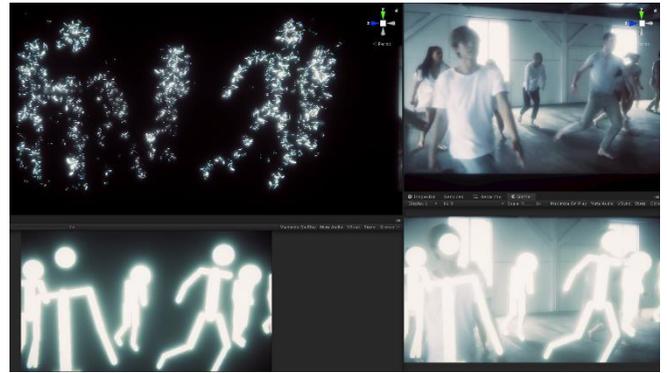

Fig. 10: Example of how stylised renders can be used with masking/detection for artistic endevours [51].  On top right of the image is a video with people and top left showcasing stylized particle system render. Bottom of image shows detection. To note: the example redition in the image is done on skeleton track only, through openpose[19] and not on perfect mask/outline of people.

## 5   Discussion

Initial implementation and experimentation of szloca's 3D position detection module is built upon OpenPose repository because of easy availability of their OpenPose Unity-Plugin.

The method is bit a dependent on camera's viewing and positioned angle, for example a camera pointed horizontally to waist height or to horizon will not work, however szloca works best in slightly tilted to the ground camera's, which most of the surveillance or any field cameras are.

Szloca method has initially been tested where you need to define the ground plane prior - to obtain 3d position of objects/person and the camera needs to be placed still, however it can further be combined with plane detection systems [32] to make it fully independent 3d position detection system , and possibly working on moving camera systems as well instead of just still.

The paper inclines more towards human tracking however the same principles can be used to get virtual data for objects, vehicles etc.

Few recent researches propose alternate methods to achieve 3D pose and position tracking [50] [9] [46], these can further be combined with Tracking and ID modules and furthermore integration with Unity. Szloca's ray approach to detect position offers a quick guerilla method to obtain 3D position from a 2D video feed and is easier to implement in game engine environments. It can also further be used to refine other approaches and vice versa. There is still uncertainty between how these work in different environments and types of videos and may or may not perform better than szloca, in terms of computation power, distant objects/people etc, - further tests and data is needed for comparisons.

# 6 Conclusions

On the context of New Media Arts, the paper discusses ways to obtain useful data to build interactivity upon while surpassing the limitations of prior technologies enabling tracking solutions over larger areas of multiple people.

The research positions a framework towards creating full fledged object/people tracking system from a single camera, including 3D world position, 3D skeleton generation, Tracking/ID and masking/classification. While looking at the state of the art in the field of computer vision and human presence tracking, the paper proposes an original method to obtain 3D position of objects from 2D video/monocular images, ie- detecting objects/people in a large area with three dimensional position data and virtual recreation.

Use cases of such technology and data are huge in every industry from human interaction systems, pedestrian behavious patterns, transportation, inventory, warehousing, sports, surveillance, monitoring to interactive gaming/arts industry where it can used to process videos, make interactive collaborative games , serious games – training, tracking and simulation, or interactive art installations without depending on the expensive hardware and their limitations for people/object tracking. It can be used, to create interactions between people, pose, environment and position-based visualizations and interactivity, and make surfaces interactive without having the need to add more tech on actual physical surfaces. Author is currently experimenting with such concepts in his projects implementing this tech. [69]

## Acknowledgment

We acknowledge all the works on which the framework of this research is build upon. Author wishes to thank Matthew Riley, Chris Barker, Playable Cities x RMIT Team, Orcha Collective and MAGI Studio at RMIT University.

Interactive Space. *Contemporary Arts and Cultures*. Retrieved from https://contemporaryarts.mit.edu/pub/dancing-with-interactive-space

## Works consulted

**Sahaj Garg** is a new media artist, creative technologist and a creative hacker experimenting around interactivity and immersive spaces. He has completed his M.S. in Animation, Games and Interactivity (MAGI) in 2020 from RMIT University and is founding member of XR club at MAGI Studios. He receoved his B.S. in Animation, Interactive Technology, Video Graphics and Special Effects from Manipal University in 2017. He often works as freelancer in augmented, virtual and mixed reality, Immersive Installations, interactivity and game development, His works have been showcased in collaborations with RMIT MAGI Expo, reAction Theatre, Playable Cities and further projects lined up. https://design.embracingearth.space/sahaj-garg/.

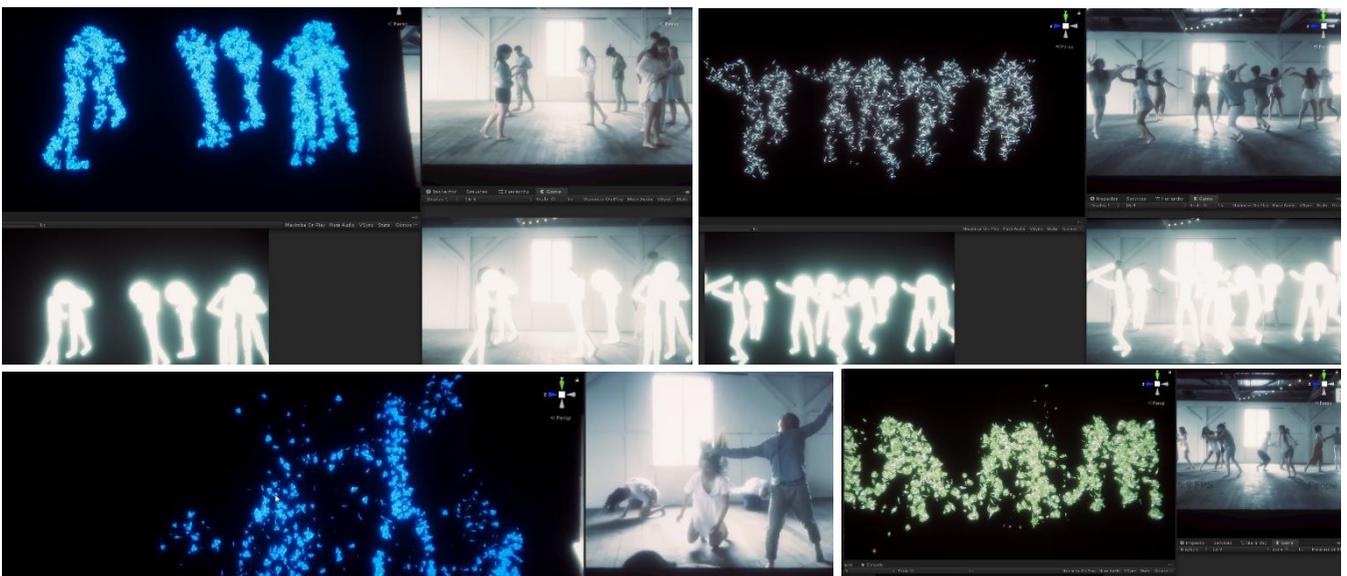

Fig. 11. Stylised Renders on detected Mask/Pose of people.

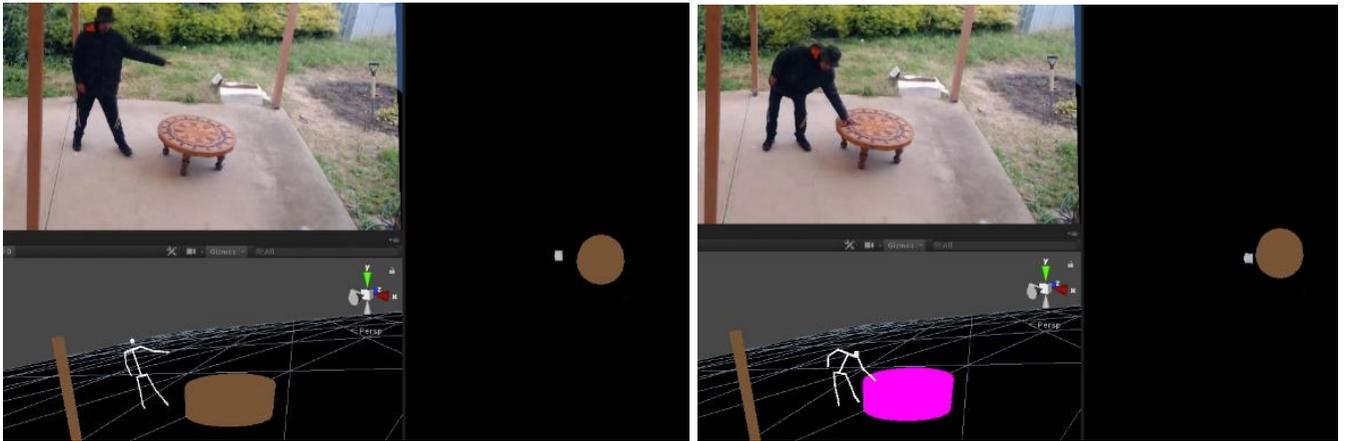

Fig. 12. Example of detecting touch over surfaces without any hardware. This enables to make huge physical objects interactive.

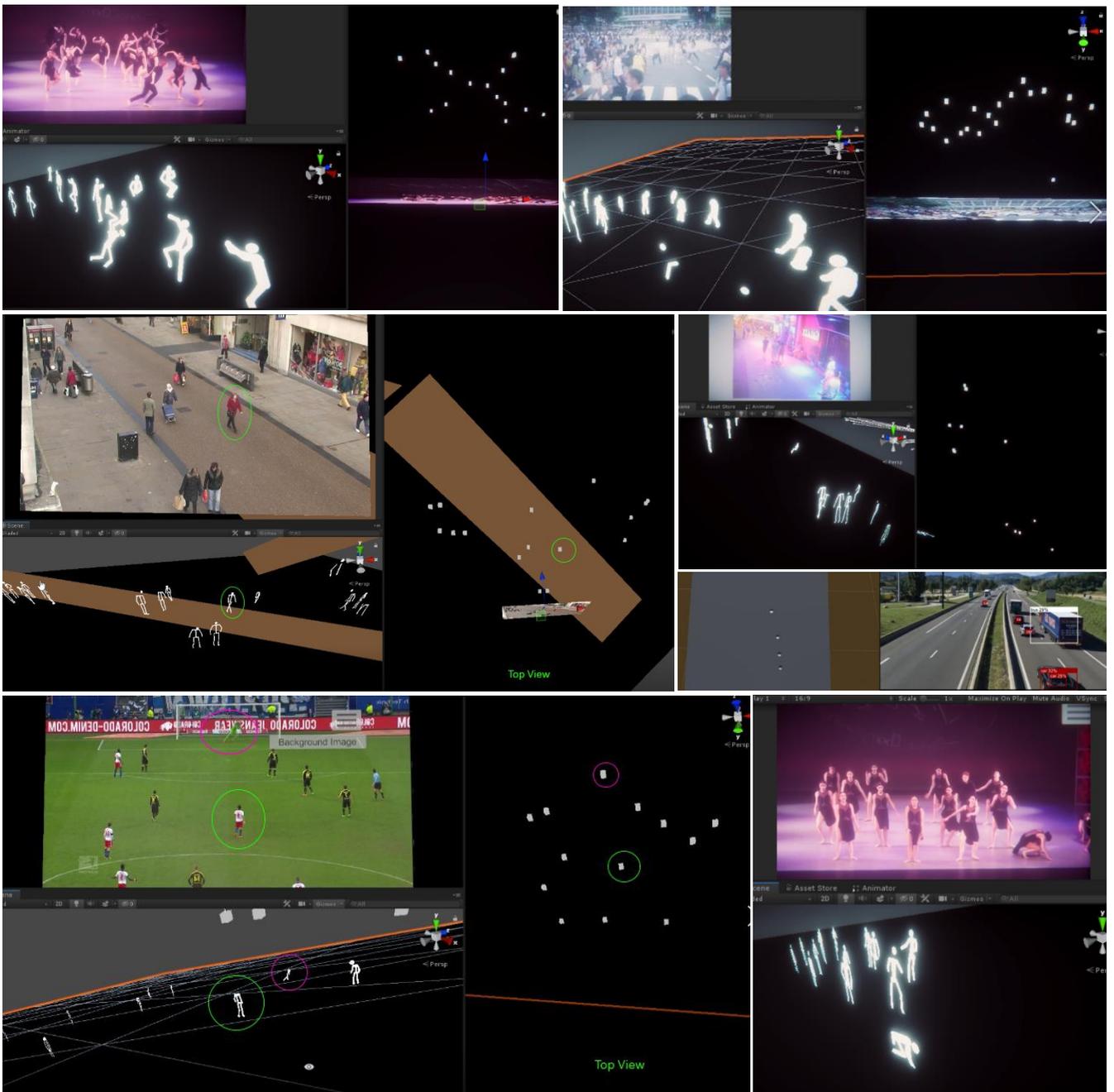

Fig. 13. SzLoca's framework in different scenarios, showing 3D position data and skeletons from monocular videos.